\documentclass[letterpaper, 10 pt, conference]{ieeeconf}  

\IEEEoverridecommandlockouts                              

\usepackage{float}
\usepackage[caption = true]{subfig}
\usepackage{url}
\usepackage{bm,bbold}
\usepackage{textcomp}
\usepackage{indentfirst}
\usepackage{xcolor}

\let\labelindent\relax
\usepackage{enumitem}
\usepackage{graphicx}
\usepackage{amsmath}
\usepackage{amssymb}
\usepackage[draft,breaklinks=true,bookmarks=false]{hyperref}
\usepackage{booktabs}

\graphicspath{{figure/}}

\begin{document}
\title{\LARGE \bf Sensor Transfer: Learning Optimal Sensor Effect Image Augmentation for Sim-to-Real Domain Adaptation}

\author{Alexandra Carlson$^{1}$, Katherine A. Skinner$^{1}$, Ram Vasudevan$^{2}$ and Matthew Johnson-Roberson$^{3}$
\thanks{$^{1}$A. Carlson and K. A. Skinner are with the Robotics Institute, University of Michigan, Ann Arbor, MI 48109, USA {\tt\small \{askc, kskin\}@umich.edu}}%
\thanks{$^{3}$R. Vasudevan is with the Department of Mechanical Engineering, University of Michigan, Ann Arbor, MI 48109, USA {\tt\small ramv@umich.edu}}
\thanks{$^{3}$M. Johnson-Roberson is with the Department of Naval Architecture and Marine Engineering, University of Michigan, Ann Arbor, MI 48109, USA {\tt\small mattjr@umich.edu}}
}
\maketitle
\begin{abstract}
\label{sec:abstract}
Performance on benchmark datasets has drastically improved with advances in deep learning. Still, cross-dataset generalization performance remains relatively low due to the domain shift that can occur between two different datasets. This domain shift is especially exaggerated between synthetic and real datasets. 
%
Significant research has been done to reduce this gap, 
specifically via modeling variation in the spatial layout of a scene, such as occlusions, and scene environmental factors, such as time of day and weather effects. 
However, few works have addressed modeling the variation in the sensor domain as a means of reducing the synthetic to real domain gap. 
The camera or sensor used to capture a dataset introduces artifacts into the image data that are unique to the sensor model, suggesting that sensor effects may also contribute to domain shift. 
To address this, we propose a learned augmentation network composed of physically-based augmentation functions. Our proposed augmentation pipeline transfers specific effects of the sensor model -- chromatic aberration, blur, exposure, noise, and color temperature -- from a real dataset to a synthetic dataset. 
We provide experiments that demonstrate that augmenting synthetic training datasets with the proposed learned augmentation framework reduces the domain gap between synthetic and real domains for object detection in urban driving scenes.

\end{abstract}


\section{Introduction}
%
%





\begin{figure}[t]
\begin{center}
   \includegraphics[width=1.0\linewidth]{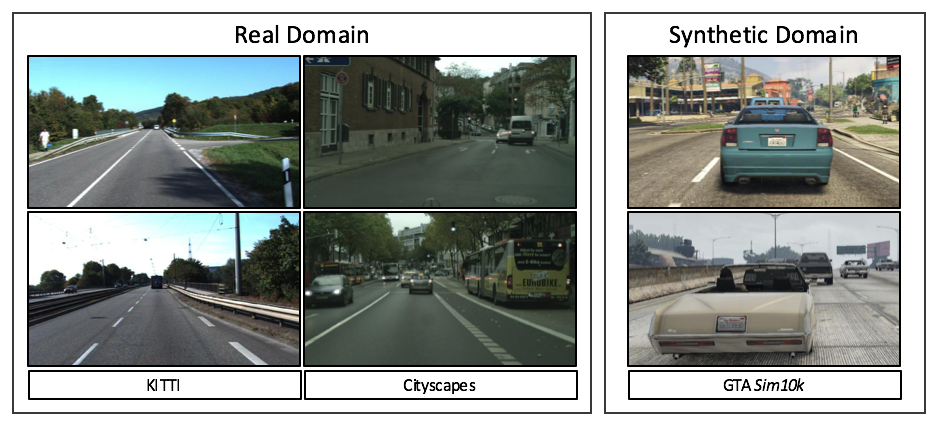}
\end{center}
   \caption{A comparison of images sampled from the real domain, KITTI Benchmark dataset (shown in the left hand column), images taken from the Cityscapes dataset (shown in the center column), and images from GTA \textit{Sim10k} dataset (shown in the right hand column). Note that each dataset has a distinct visual style, specifically differing color cast, brightness, and blur.}
\label{fig:st_intro_fig}
\end{figure}

Synthetic datasets are designed to contain numerous spatial and environmental features that are found in the real domain: images captured during different times of day, in various weather conditions, and in structured urban environments.
However, in spite of these shared features and high levels of photorealism, images from synthetic datasets are noticeably stylistically distinct from real images. Figure~\ref{fig:st_intro_fig} shows a side-by-side comparison of two of widely-used real benchmark vehicle datasets, KITTI~\cite{kittiobject,KITTI13}, Cityscapes~\cite{Cityscapes}, and a state-of-the-art synthetic dataset, GTA \textit{Sim10k}~\cite{Richter_2016_ECCV,johnson2017driving}. These differences can be quantified; a performance drop is observed between training and testing deep neural networks (DNNs) between the synthetic and real domains~\cite{johnson2017driving}. This suggests that real and synthetic datasets differ in their global pixel statistics.
%
%
%
Domain adaptation methods attempt to minimize such dissimilarities between synthetic and real datasets that result from an uneven representation of visual information in one domain compared to the other. 
Recent domain adaptation research has focused on learning salient visual features from real data -- specifically scene lighting, scene background, weather, and occlusions -- using generative adversarial frameworks in an effort to better model the representation of these visual elements in synthetic training sets~\cite{zhang2017image,veeravasarapu2017adversarially,sakaridis2018model}. 
%
However, little work has focused on modelling realistic, physically-based augmentations of synthetic data. 
Carlson et al.~\cite{carlson2018modeling} demonstrate that randomizing across the sensor domain significantly improves performance over standard augmentation techniques. 
The information loss that results from the interaction between the camera model and lighting in the environment is not generally modelled in rendering engines, despite the fact that it can greatly influence the pixel-level artifacts, distortions, and dynamic range, and thus the global visual style induced in each image ~\cite{grossberg2004modelingCRF,couzinie2013learning_modelingblur,foi2008practical,andreopoulos2012sensor,song2015sun,dodge2016understanding,doersch2015unsupervised}.  
%

In this study, we build upon~\cite{carlson2018modeling} to work towards closing the gap between real and synthetic data domains. 
We propose a novel learning framework that performs \textit{sensor transfer} on synthetic data. That is, the network learns to transfer the real sensor effect domain -- blur, exposure, noise, color cast, and chromatic aberration -- to synthetic images via a generative augmentation network. 
We demonstrate that augmenting relatively small labeled datasets using \textit{sensor transfer} generates more robust and generalizable training datasets that improve the performance of DNNs for object detection and semantic segmentation tasks in urban driving scenes for both real and synthetic visual domains.


This paper is organized as follows: Section~\ref{sec:background} presents related background work; section~\ref{sec:Methods} details the proposed \textit{sensor transfer} learning framework; section ~\ref{sec:Experiments} describes experiments and discusses results of these experiments and section~\ref{sec:concl} concludes the paper. Code will be made publicly available.
%

\section{Related work}
\label{sec:background}
%
%
Our work focuses on augmenting the training data directly so that it can be applied to any task or input into any deep neural network regardless of the architecture. 
Zhang et al. 2017~\cite{zhang2017physically} demonstrate that the level of photorealism of the synthetic training data directly impacts the robustness and performance of the deep learning algorithm when tested on real data across a variety of computer vision tasks. However, it remains unclear what features of real data are necessary for this performance gain, or what parts of rendering pipelines should be modified to bridge the synthetic to real domain gap. 
Much work in the fields of data augmentation and learned rendering pipelines have proposed methods that shed light on  this topic, and are summarized below. 
\subsection{Domain Randomization} 
Recent work on domain randomization seeks to bridge the sim-to-real domain gap by generating synthetic data that has sufficient random variation over scene factors and rendering parameters such that the real data falls into this range of variation, even if the rendered data does not appear photorealistic. Such scene factors include such as textures, occlusion levels, scene lighting, camera field of view, and uniform noise, and have been applied to vision tasks in complex indoor and outdoor environments~\cite{tobin2017domain,tremblay2018training}. 
The drawback of these techniques is that they only work if they sample the visual parameters spaces finely enough, and create a large enough dataset from a broad enough range of visual distortions to encompass the variation observed in real data. This can result in intractably large datasets that require significant training time for a deep learning algorithm.
While we also aim to achieve robustness via an augmentation framework, we can use smaller datasets to achieve state-of-the-art performance because our method is learning how to augment synthetic data with salient visual information that exists in real data.
Note that, because our work focuses on image augmentation outside of the rendering pipeline, it could be used in addition to domain randomization techniques.
%
\subsection{Optimizing Augmentation}
In contrast to domain randomization, task-dependent techniques have been proposed to achieve more efficient data augmentation by learning the type and number of image augmentations that are important for algorithm performance.
State-of-the-art methods~\cite{paulin2014transformation,cubuk2018autoaugment,lemley2017smart} in this area treat data augmentation as a form of network regularization, selecting a subset of augmentations that optimize algorithm performance for a given dataset and task as the algorithm is being trained.
Unlike these methods, we propose that data augmentation can function as a domain adaptation method. Our learning framework is task-independent, and uses physically based augmentation techniques to investigate the visual degrees of freedom (defined by physically-based models) necessary for optimizing network performance from the synthetic to real domain. 


\subsection{Image-to-Image Translation for Domain Adaptation}
Impressive advances have been made in both paired and unpaired image-to-image translation ~\cite{CycleGAN2017,johnson2016perceptual,zhu2016generative,isola2017image,huang2018multimodal,liu2017unsupervised} to bridge a variety of domain gaps, including season-to-season, night-to-day, and sim-to-real.
However, image-to-image translation performed between image sets with complex, varied environments often introduces unrealistic distortion artifacts into the underlying structure of the scene. This can yield poor performance for visual tasks such as object detection and semantic segmentation~\cite{dundar2018domain}. 
In contrast, the proposed method does not alter the spatial information in the scene, and instead translates images from one domain to another constrained by physically-based image augmentation. 

\subsection{Learned Rendering Pipelines for Domain Adaptation}
Several studies have proposed unsupervised, generative learning frameworks that either take the place of a standard rendering engine~\cite{veeravasarapu2017adversarially} or complement the rendering engine via post-processing~\cite{sixt2016rendergan,simgan,huangexpecting} in order to model relevant visual information directly from real images with no dependency on a specific task framework.  
Both ~\cite{veeravasarapu2017adversarially} and ~\cite{huangexpecting} are applied to complex outdoor image datasets, but are designed to learn distributions over simpler spatial features in real images, specifically scene geometry. Other methods, such as ~\cite{sixt2016rendergan,simgan}, attempt to learn low-level pixel features. However, they are only applied to image sets that are homogeneously structured and low resolution. This may be due to the sensitivity of training adversarial frameworks. 
Our work focuses specifically on modeling the camera and image processing pipeline rather than scene elements or environmental factors that are specific to a given task. Our method can be applied to high resolution images of complex scenes.
%
%
\subsection{Impact of Sensor Effects on Deep Learning} 
Recent work has demonstrated that elements of the image formation and processing pipeline can have a large impact upon learned representation for deep neural networks across a variety of vision tasks~\cite{Kanan2012,diamond2017dirty,dodge2016understanding,doersch2015unsupervised}. 
The majority of methods propose learning techniques that remove these effects from images~\cite{diamond2017dirty}. As many of these sensor effects can lead to loss of information, correcting for them is non-trivial, potentially unstable, and may result in the hallucination of visual structure in the restored image. 
In contrast, Carlson et al.~\cite{carlson2018modeling} demonstrate that significant performance boosts can be achieved by augmenting images using physically-based, sensor effect domain randomization.
However, their method requires hand-tuning/evaluation of the visual quality of image augmentation. This human-in-the-loop dependence is inefficient and difficult to scale for large synthetic datasets, and the evaluated visual image quality is subjective.
Rather than removing these effects, randomly adding them in, or manually adding them in via human-in-the-loop, our method learns the the style of sensor effects from real data and transfers this \textit{sensor style} to synthetic images to bridge the synthetic-to-real domain gap. 


\section{Methods}
\label{sec:Methods}
%
\begin{figure*}[ht]
\begin{center}
\includegraphics[width=0.8\linewidth]{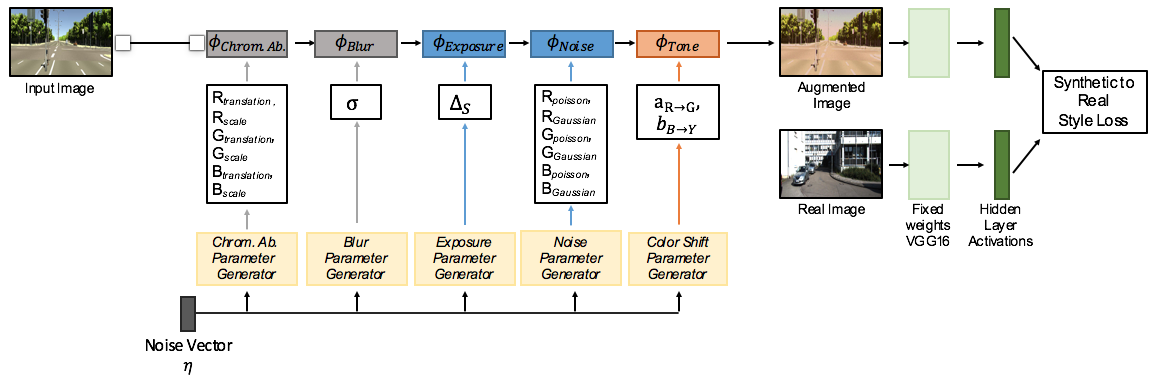}
\caption{The schematic of the proposed sensor transfer network structure. The style loss trains the sensor effect parameter generators (represented as the yellow boxes) to select parameters that transform the input synthetic images based upon how the sensor effect augmentation functions alter the style of the real data domain. This effectively transfers 'sensor style' of the target dataset to the source dataset.}
\label{fig:STnet}
\end{center}
\end{figure*}
The objective of the sensor transfer network is to learn the the optimal set of augmentations that transfer sensor effects from a real dataset to a synthetic dataset. 
Our complete Sensor Transfer Network is shown in Figure~\ref{fig:STnet}.

\vspace{-4mm}
\subsection{Sensor Effect Augmentation Pipeline}

We adopt the sensor effect augmentation pipeline from ~\cite{carlson2018modeling}. This is the backbone of the Sensor Transfer Network. Refer to~\cite{carlson2018modeling} for a detailed discussion of each function and its relationship to the image formation process in a camera. We briefly describe each sensor effect augmentation function below for completeness.
The sensor effect augmentation pipeline is a composition of chromatic aberration, Gaussian blur, exposure, pixel-sensor noise, and post-processing color balance augmentation functions: 
\begin{align}
I_{aug.} = f_{color}(f_{noise}(f_{exposure}(f_{blur}(f_{chrom.ab.}(I)))))
\label{eq:aug-pipeline-eq}
\end{align}

\noindent\textbf{Chromatic Aberration}\\ 
To model lateral chromatic aberration, we apply translations $(t_{x},t_{y})$ in 2D pixel space to each of the color channels of an image. To model longitudinal chromatic aberration, we scale the green color channel relative to the red and blue channels of an image by a value $S$. We combine these parameters into an affine transformation on each pixel in color channel of the image. The augmentation parameters learned for this augmentation function are $S$, the red channel translations $R_{x}$ and $R_{y}$, the green channel translations $G_{x}$ and $G_{y}$, and the blue channel translations $B_{x}$ and $B_{y}$.\\
%

\noindent\textbf{Blur} \\
We implement out-of-focus blur, which is modeled by convolving the image with a Gaussian filter~\cite{cheong2015fast}.
We fix the window size of the kernel to 9.0. The augmentation parameter learned for this augmentation function is the standard deviation $\sigma$ of the kernel.\\

\noindent\textbf{Exposure}\\ 
We implement the exposure density function developed in~\cite{exp1, exp2}:
\begin{align}
I = f(S) = \frac{255}{1 + e^{-A \times S}}
\label{eq:exposure}
\end{align}
\noindent where $I$ is image intensity, $S$ models the incoming light intensity, and $A$ is a constant value that describes image contrast. We set $A$ to 0.85. This model is used to re-expose an image as follows:
\begin{align}
S' = f^{-1}(I) +  \Delta S
\label{eq:exposureds}
\end{align}
\begin{align}
I_{exp} = f(S')
\label{eq:exposureimg}
\end{align}
\noindent The augmentation parameters learned for this augmentation function are $\Delta S$ to model changing exposure, where a positive $\Delta S$ relates to increasing the exposure, and a negative value indicates decreasing exposure.\\

\noindent\textbf{Noise}\\
We use the Poisson-Gaussian noise model proposed in~\cite{foi2008practical}: 
\begin{align}
I_{noise}(x,y)=I(x,y)+\eta_{poiss}(I(x,y))+\eta_{gauss}
\label{eq:noise1}
\end{align}
where $I(x,y)$ is the ground truth image at pixel location $(x,y)$, $\eta_{poiss}$ is the signal-dependent poisson noise, and $\eta_{gauss}$ is the signal-independent gaussian noise.
The augmentation parameters learned for this augmentation function are the $\eta_{poiss}$ and $\eta_{gauss}$ for each color channel, for a total of six parameters.\\

\noindent\textbf{Post-processing}\\
We model post-processing techniques done by cameras, such as white balancing or gamma transformation, by performing linear translations in LAB color space~\cite{hunter1948accuracyLAB,annadurai2007fundamentals}. 
The augmentation parameters learned for this augmentation function the are translations in the a (red-green)and b (blue-yellow) channels in normalized LAB space.
%
\begin{figure}
\begin{center}
\includegraphics[width=1.0\linewidth]{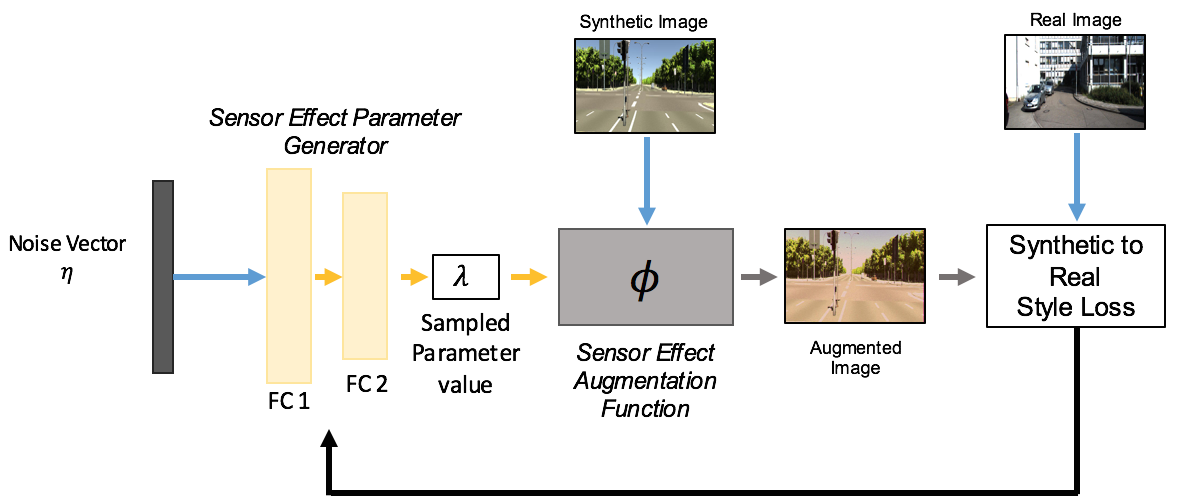}
\caption{A detailed schematic of how the training process occurs for a single sensor effect function. A 200 dimensional uniform noise vector (sampled from the range -1 to 1) is generated for a given input synthetic image. The uniform noise vector is input into the fully connected neural network that constitutes the parameter generator, which outputs sampled value(s) for  the respective sensor effect augmentation function. The sampled parameter value(s) and the input synthetic image are fed into the sensor effect augmentation function, which outputs an augmented synthetic image. The style loss is calculated between the augmented synthetic image and a real image. This style loss is then backpropagated through the augmentation functions to train the parameter generator to select parameters that reduce the style differences between the real and augmented synthetic images.\vspace{-6mm}}
\label{fig:training_step_diagram}
\end{center}
\end{figure}

\subsection{Training the Sensor Transfer Network}
A high-level overview of a single training iteration for a single sensor effect is given in Figure~\ref{fig:training_step_diagram}.
Each sensor effect augmentation function has its own parameter generator network. The training objective for each of these networks is to learn the distribution over its respective augmentation parameter(s) based upon real data. Each generator network is a two-layer, fully connected neural network. 
The following steps are required to perform a single training iteration of the Sensor Transfer Network using a single synthetic image. First, a 200 dimensional uniform noise vector, $\eta$, is generated and paired with the input synthetic image. The noise vector $\eta$ is input into each separate generator network.  Each generator network consists of two fully connected layers that together project $\eta$ into its respective sensor effect parameter space. For example, the blur parameter generator will map the $\eta$ to a value in the $\sigma$ parameter space. 
The output sampled parameters, paired with the input synthetic image are then input into the augmentation pipeline, which outputs an augmented synthetic image. 
This augmented synthetic image is then paired with a real image, both of which are input to the loss function.  \\
We employ a loss function similar to the one used in Johnson et. al~\cite{johnson2016perceptual}. We assume that the layers of the VGG-16 network~\cite{vgg16} trained on ImageNet~\cite{deng2009imagenet} encode relevant style information for salient objects
We fix the weights of the pretrained VGG-16 network, and use it to project real and augmented synthetic images into the hidden layer feature spaces. 
We calculate the style loss, given in Eqn.~\ref{styleloss}, and use this as the training signal for the parameter generators.
\begin{align}
L_{style}(y,\hat{y})= \Sigma_{j}\Vert G^{\theta}_{j}(y) - G^{\theta}_{j}(\hat{y}) \Vert ^{2}_{Frobinius}
\label{styleloss}
\end{align}
\vspace{-2mm}
In the above equation, $y$ is a real image batch, $\hat{y}$ is an augmented synthetic image batch, $G^{\theta}_{j}(y)$ is the Gram matrix of the feature map $\theta(y)$ of hidden layer $j$ of the pretrained VGG-16 network, and $G^{\theta}_{j}(\hat{y})$ is the corresponding quantity for augmented synthetic images. Through performance-based ablation studies, we found that $j$ = 10 gives the best performance, so the style loss is calculated for the first ten layers of VGG-16.
Once calculated, the style loss is backpropagated through the sensor effect augmentation functions to train the sensor effect parameter generators. 
The above process is repeated with images from the synthetic and real datasets until the style loss has converged.

We train the sensor effects generators concurrently to learn the joint probability distribution over the sensor effect parameters. This is done to capture the dependencies that exist between these effects in a real camera. 
Once training is complete, we can fix the weights of the parameter generators, and use them to sample learned parameters to augment synthetic images. Table~\ref{table:ST-hyperparameters} shows the statistics of the learned distributions for sensor effect parameters of different real datasets. See Section~\ref{sec:Experiments} for analysis and discussion of the learned parameters. 
Note that style loss was chosen because it is independent of spatial structure of an image.
In effect, the augmentation parameter generators learn to sample the distributions of sensor effects in real data as constrained by the style of the real image domain. 

\begin{figure*}
\begin{center}
\includegraphics[width=0.9\linewidth]{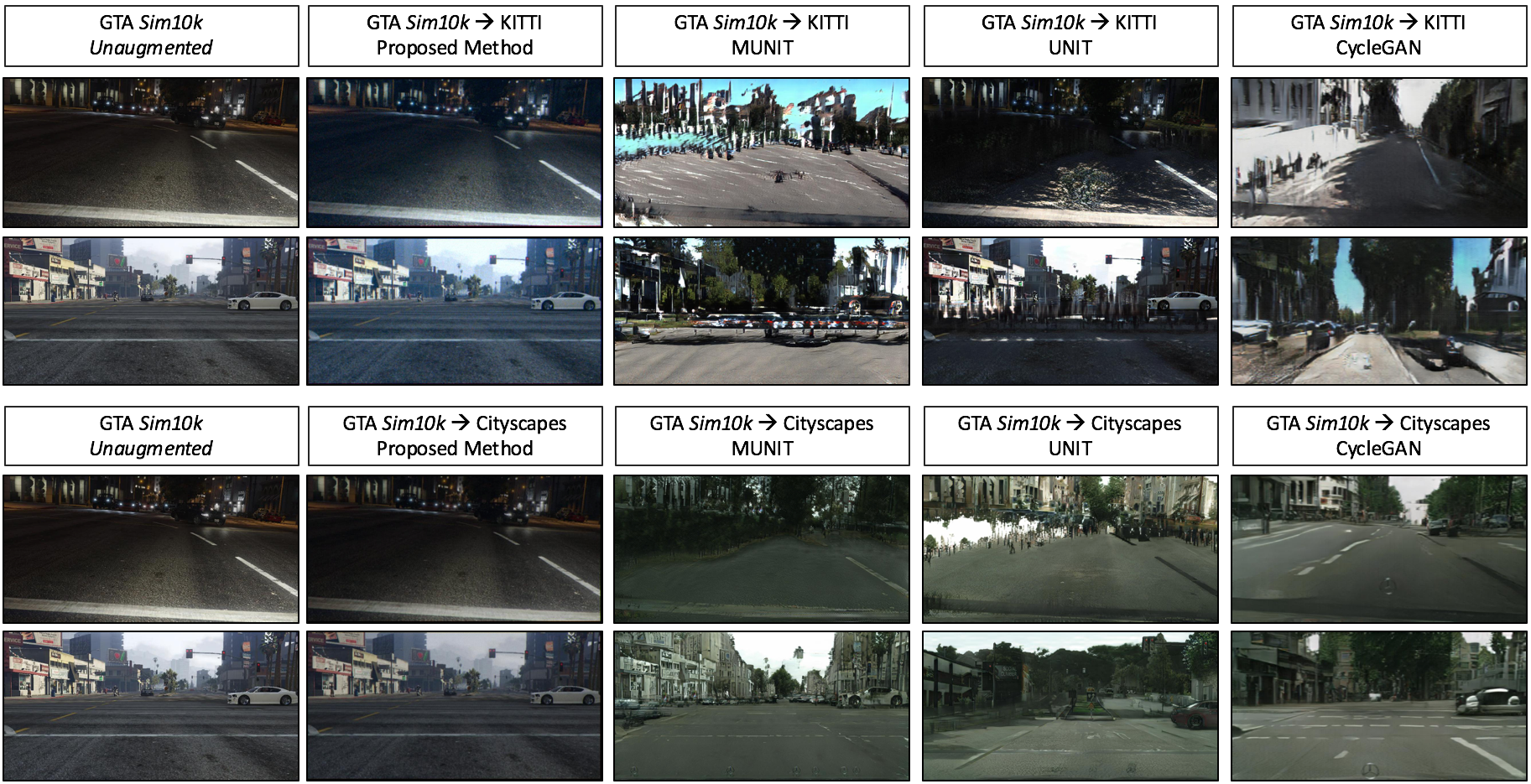}
\end{center}
\caption{Qualitative comparison of unaugmented GTA\textit{Sim10k} in the first column, Sensor Transfer augmented GTA\textit{Sim10k} images in the second column,  MUNIT augmented GTA\textit{Sim10k} in the third colum, UNIT-augmented GTA\textit{Sim10k} in the fourth column, and CycleGAN-augmented GTA\textit{Sim10k} in the last column. The first two rows are GTA\textit{Sim10k} translated to the KITTI domain, and the second two rows are GTA\textit{Sim10k} translated into the Cityscapes domain. Note that, for the Sensor Transfer augmented images, the primary sensor effect transferred in GTA\textit{Sim10k}$\rightarrow$Cityscapes augmentation is decreased exposure, whereas the primary sensor effects transferred in GTA\textit{Sim10k}$\rightarrow$KITTI augmentation is a blueish hue and increased exposure. In comparison, images augmented using the image-to-image translation networks lose a significant amount of spatial information. These methods also cannot handle night time images as well as the proposed method.}
\label{fig:qualitative-results}
\end{figure*}

\section{Experiments}
\label{sec:Experiments}
\subsection{Experimental Setup}
To verify that the proposed method can transfer the sensor effects of different datasets, we train Sensor Transfer Networks using the following synthetic and real benchmark datasets: 
GTA\textit{Sim10k}~\cite{johnson2017driving} is comprised of 10,000 highly photorealistic synthetic images collected from the Grand Theft Auto (GTA) rendering engine. It captures different weather conditions and time of day. 
The Cityscapes~\cite{Cityscapes} training image set is comprised of 2975 real images collected in over 50 cities across Germany. 
The KITTI training set~\cite{kittiobject} is comprised of 7481 real images collected in Karlshue, Germany. We train a Sensor Transfer Network to transfer the sensor style of the KITTI training set to GTA\textit{Sim10k}, which is referred to as GTA\textit{Sim10k}$\rightarrow$KITTI. We also train a Sensor Transfer Network to transfer the sensor style of the Cityscapes training set to GTA\textit{Sim10k}, which is referred to as GTA\textit{Sim10k}$\rightarrow$Cityscapes. To train each Sensor Transfer Network, we use a batch size of 1 and learning rate of $2e^{-5}$. We trained each network for 4 epochs.
For all experiments, we compare our results to the Sensor Effect Domain Randomization~\cite{carlson2018modeling} of GTA\textit{Sim10k} as a baseline measure to ensure that the transfer of effects is viable over sampling. To generate the Sensor Effect Domain Randomization augmentations, we used the same human-selected parameter ranges as in~\cite{carlson2018modeling}. To benchmark our method against other, image-based domain adaptation methods, we use the state-of-the-art image-to-image translation methods CycleGAN~\cite{CycleGAN2017}, UNIT~\cite{liu2017unsupervised}, and MUNIT~\cite{huang2018multimodal} as additional baseline measures. Each of the CycleGAN, UNIT, and MUNIT image-to-image translation networks were trained to transfer GTA\textit{Sim10k} to Cityscapes, and separately to transfer GTA\textit{Sim10k} to KITTI. Each network was trained using either the default hyperparameters provided in the respective paper(s) or until the networks converged.
%
%

\subsection{Evaluation of Learned Sensor Effect Augmentations}
Qualitatively, from observing Figure~\ref{fig:st_intro_fig}, KITTI images feature more pronounced visual distortions due to blur, over-exposure, and a blue color tone. Cityscapes, on the other hand, has a more under-exposed, darker visual style. 

\begin{table*}
\begin{center}
\caption{Learned sensor effect parameters for  GTA\textit{Sim10k}$\rightarrow$Cityscapes and GTA\textit{Sim10k}$\rightarrow$KITTI, and the Sensor Effect Domain Randomization parameters from Carlson et al.~\cite{carlson2018modeling}. Note that for the Sensor transferred parameters in the first two rows, the mean and standard deviation of each sensor effect parameter value is given in the convention $\mu\pm\sigma$. For the Carlson et al.~\cite{carlson2018modeling} Sensor Effect Domain Randomization parameters, given in the final row, the minimum and maximum of the human selected range is provided. Quantitatively, the GTA\textit{Sim10k}$\rightarrow$KITTI increases image exposure, adds chromatic aberration, noise,  and adds a blue color cast. For GTA\textit{Sim10k}$\rightarrow$Cityscapes, image exposure is decreased, adds chromatic aberration, a higher level of noise is added, and slight yellow-blue color cast is applied. }
\label{table:ST-hyperparameters}
\begin{tabular}{c}
\begin{tabular}{|c|c|c|c|c|}
\multicolumn{5}{c}{Proposed Method GTA\textit{Sim10k}$\rightarrow$Cityscapes Sensor Effect Parameters}  \\
\hline
{\scriptsize Chrom. Ab.} & {\scriptsize Blur} & {\scriptsize Exposure} & {\scriptsize Noise} & {\scriptsize Post-processing} \\
\hline\hline
$G_{scale}$: $0.999\pm2.398e^-5$ & $\sigma$: $0.718\pm1.34e^-13$   & $\Delta S$: $-0.273\pm0.0249$ & $R_{gauss.}$: $1.0e^-6\pm1.382e^-18$  & $a$:$-0.002\pm5.239e^-4$ \\
$R_{tx}$: $0.004\pm6.221e^-5$    &                                 &                               & $R_{poiss.}$: $1.0e^-6\pm1.382e^-18$  & $b$: $-0.0116\pm4.727e^-4$ \\
$R_{ty}$: $0.007\pm5.511e^-5$    &                                 &                               & $G_{gauss.}$: $5.41\pm4.249e^-4$     &  \\
$G_{tx}$: $0.005\pm1.111e^-5$    &                                 &                               & $G_{poiss.}$: $1.15e^-2\pm7.913e^-5$ &  \\
$G_{ty}$: $0.006\pm4.718e^-5$    &                                 &                               & $B_{gauss.}$: $1.0e^-6\pm1.382e^-18$  &  \\
$B_{tx}$: $0.006\pm5.793e^-5$    &                                 &                               & $B_{poiss.}$: $6.8e^-4\pm4.608e^-6$ &  \\
$B_{ty}$: $-5.052\pm1.16e^-4$     &                                 &                              &                                       &  \\
\hline
\end{tabular}
\\
\\
\begin{tabular}{|c|c|c|c|c|}
\multicolumn{5}{c}{Proposed Method GTA\textit{Sim10k}$\rightarrow$KITTI Sensor Effect Parameters}  \\
\hline
{\scriptsize Chrom. Ab.} & {\scriptsize Blur} & {\scriptsize Exposure} & {\scriptsize Noise} & {\scriptsize Post-processing} \\
\hline
$G_{scale}$: $1.001\pm6.425e^-5$  & $\sigma$: $0.941\pm5.173e^-7$   & $\Delta S$: $0.0823\pm0.003$  & $R_{gauss.}$: $9.5e^-3\pm3.713e^-4$ & $a$:$-0.0131\pm5.426e^-4$ \\
$R_{tx}$: $1.134e^-4\pm9.416e^-5$ &                                 &                               & $R_{poiss.}$: $3.07e^-2\pm1.295e^-3$ & $b$: $-0.0882\pm3.25e^-3$ \\
$R_{ty}$: $-0.0013\pm6.874e^-5$   &                                 &                               & $G_{gauss.}$: $4.5e^-3\pm2.005e^-4$ &  \\
$G_{tx}$: $-4.67e^-4\pm5.65e^-5$  &                                 &                               & $G_{poiss.}$: $2.62e^-2\pm1.111e^-3$ &  \\
$G_{ty}$: $-0.0014\pm7.228e^-5$   &                                 &                               & $B_{gauss.}$: $2.65e^-2\pm1.111e^-3$ &  \\
$B_{tx}$: $-0.003\pm1.245e^-4$    &                                 &                               & $B_{poiss.}$: $4.47e^-2\pm1.187e^-3$ &  \\
$B_{ty}$: $-5.16e^-5\pm1.096e^-4$ &                                 &                               &                                    &  \\
\hline
\end{tabular}
\\
\\
\begin{tabular}{|c|c|c|c|c|}
\multicolumn{5}{c}{Carlson et al.~\cite{carlson2018modeling} Sensor Effect Domain Randomization Parameters}  \\
\hline
\hline
$G_{scale}$: 0.998-1.002 & $\kappa_{size}$: 3-11 & $\Delta S$: -0.6-1.2 & $R_{gauss.}$: 0.00-0.05 & $a$: -10.0-10.0 \\
$R_{tx}$: -0.003-0.003   & $\sigma$: 0.0-3.0     &                      & $G_{gauss.}$: 0.00-0.05 & $b$: -10.0-10.0 \\
$R_{ty}$: -0.003-0.003   &                       &                      & $B_{gauss.}$: 0.00-0.05 &  \\
$G_{tx}$: -0.003-0.003   &                       &                      & $R_{poiss.}$: 0.00-0.05 &  \\
$G_{ty}$: -0.003-0.003   &                       &                      & $G_{poiss.}$: 0.00-0.05 &  \\
$B_{tx}$: -0.003-0.003   &                       &                      & $B_{poiss.}$: 0.00-0.05 &  \\
$B_{ty}$: -0.003-0.003   &                       &                      &                         &  \\
\hline
\end{tabular}

\end{tabular}
\end{center}
\end{table*}
Figure~\ref{fig:qualitative-results}
shows examples of unaugmented GTA\textit{Sim10k} images in comparison to  those same images augmented by the proposed Sensor Transfer network and baseline image-to-image translation networks.
When compared to Figure~\ref{fig:st_intro_fig}, it does appear that, for both the sensor transfer of GTA\textit{Sim10k}$\rightarrow$KITTI and GTA\textit{Sim10k}$\rightarrow$Cityscapes, realistic aspects of exposure, noise, and color cast are transferred to GTA\textit{Sim10k}. 
The statistics of the learned parameter values are given in Table~\ref{table:ST-hyperparameters}. In general the selected parameter values generate augmented synthetic images with style that matches the real datasets.
We hypothesize that the color shift for GTA\textit{Sim10k}$\rightarrow$Cityscapes is not as strong as GTA\textit{Sim10k}$\rightarrow$KITTI because there is a more even distribution of sky and buildings in Cityscapes, where as KITTI has a significant number of instances of sky. Interestingly, the blur parameter, $\sigma$, did not converge and was pushed towards zero for both GTA\textit{Sim10k}$\rightarrow$Cityscapes and GTA\textit{Sim10k}$\rightarrow$KITTI. This suggests that Gaussian blur does not match the blur captured by style of real images. Further research could consider more accurate models of blur, such as motion blur. 
\subsection{Impact of Learned Sensor Transformation on Object Detection for Benchmark datasets}


\setlength{\tabcolsep}{1.0pt}
\begin{table}[h]
\begin{center}
\caption{Results of the sensor effects augmentations on Faster R-CNN object detection performance. The percent change for  CycleGAN~\cite{CycleGAN2017}, UNIT~\cite{liu2017unsupervised}, MUNIT~\cite{huang2018multimodal}, the Carlson et al.~\cite{carlson2018modeling} and proposed method are calculated relative to the full, unaugmented baseline datasets.}
\label{table:max_objdet_perf}
\begin{tabular}{c}
\begin{tabular}{|l|lllllll|}
\multicolumn{3}{c}{Training Dataset} & \multicolumn{5}{c}{Tested on KITTI}   \\
\hline
Augmentation Method                                &&& $AP_{Car}$ &&&& Gain \\
\hline
Baseline                                           &&& 51.01 &&& & \textemdash\\  
\hline
CycleGAN~\cite{CycleGAN2017}                &&& 48.75 &&& $\downarrow$& -2.25 \\
UNIT~\cite{liu2017unsupervised}             &&& 51.21 &&& $\uparrow$& 0.21 \\
MUNIT~\cite{huang2018multimodal}            &&& 45.50 &&& $\downarrow$& -5.51 \\
\hline
\textit{Carlson et al.~\cite{carlson2018modeling}} &&& 48.94 &&& $\downarrow$& -2.07 \\ 
Proposed Method                                    &&& \textbf{52.67} &&& $\uparrow$  & \textbf{+1.66}\\

\hline
\end{tabular}
\\ \\ \\
\begin{tabular}{|l|lllllll|}
\multicolumn{3}{c}{Training Dataset} & \multicolumn{5}{c}{Tested on Cityscapes}  \\
\hline
Augmentation Method                                 &&& $AP_{Car}$    &&& & Gain \\
\hline
Baseline                                                 &&& 30.13         &&& &\textemdash \\  
\hline
CycleGAN~\cite{CycleGAN2017}                       &&& 29.30 &&& $\downarrow$& -0.83 \\
UNIT~\cite{liu2017unsupervised}                    &&& 28.05 &&& $\downarrow$& -2.08 \\
MUNIT~\cite{huang2018multimodal}                   &&& 26.20 &&& $\downarrow$& -3.93 \\
\hline
\textit{Carlson et al.~\cite{carlson2018modeling}}       &&& 34.89         &&& $\uparrow$ & +4.76 \\ 
Proposed Method                                          &&&\textbf{35.48} &&& $\uparrow$ & \textbf{+5.35}\\

\hline
\end{tabular}
\end{tabular}
\end{center}
\end{table}
To evaluate if the Sensor Transfer Network is adding in salient visual information for vision tasks in the real image domain, we train an object detection neural network on the unaugmented and augmented synthetic data and evaluate the performance of the object detection network on the real data domains, KITTI and Cityscapes. 
We chose to use Faster R-CNN as our base network for 2D object detection~\cite{ren2015fasterRCNN}. 
Faster R-CNN achieves relatively high performance on the KITTI benchmark dataset. 
Many state-of-the-art object detection networks that improve upon these results still use Faster R-CNN as their base architecture.

We compare Faster R-CNN networks trained on the proposed method to Faster R-CNN networks trained on unaugmented GTA\textit{Sim10k}, GTA\textit{Sim10k} augmented using the Sensor Transfer Domain Randomization from Carlson et al., GTA\textit{Sim10k} augmented using CycleGAN, GTA\textit{Sim10k} augmented using UNIT, and GTA\textit{Sim10k} augmented using MUNIT. 
To create augmented training datasets, we combine the unaugmented GTA\textit{Sim10k} with varying amounts of augmented GTA\textit{Sim10k} data.
For all datasets, both augmented and unaugmented, we trained each Faster R-CNN network for 10 epochs using two Titan X Pascal GPUs in order to control for potential confounds between performance and training time. We evaluate the Faster R-CNN networks on either the KITTI training dataset or the Cityscapes training dataset depending on the Sensor Transfer Network used for training dataset augmentation.
Each dataset is converted into Pascal VOC 2012 format to standardize training and evaluation, and performance values are the VOC AP50 reported for the car class~\cite{pascal-voc-2012}.

Table~\ref{table:max_objdet_perf} shows the object detection results for the proposed method in comparison to the image-to-image translation and domain randomization baselines. In general, the addition of sensor effect augmentations has a positive boost on Faster R-CNN performance for training on GTA\textit{Sim10k} and testing on Cityscapes. Our proposed method, for both GTA\textit{Sim10k}$\rightarrow$Cityscapes and GTA\textit{Sim10k}$\rightarrow$KITTI, achieves the best performance over both the baseline and Sensor Effect Domain Randomization. 

\begin{figure}
\begin{center}
\includegraphics[width=1.0\linewidth]{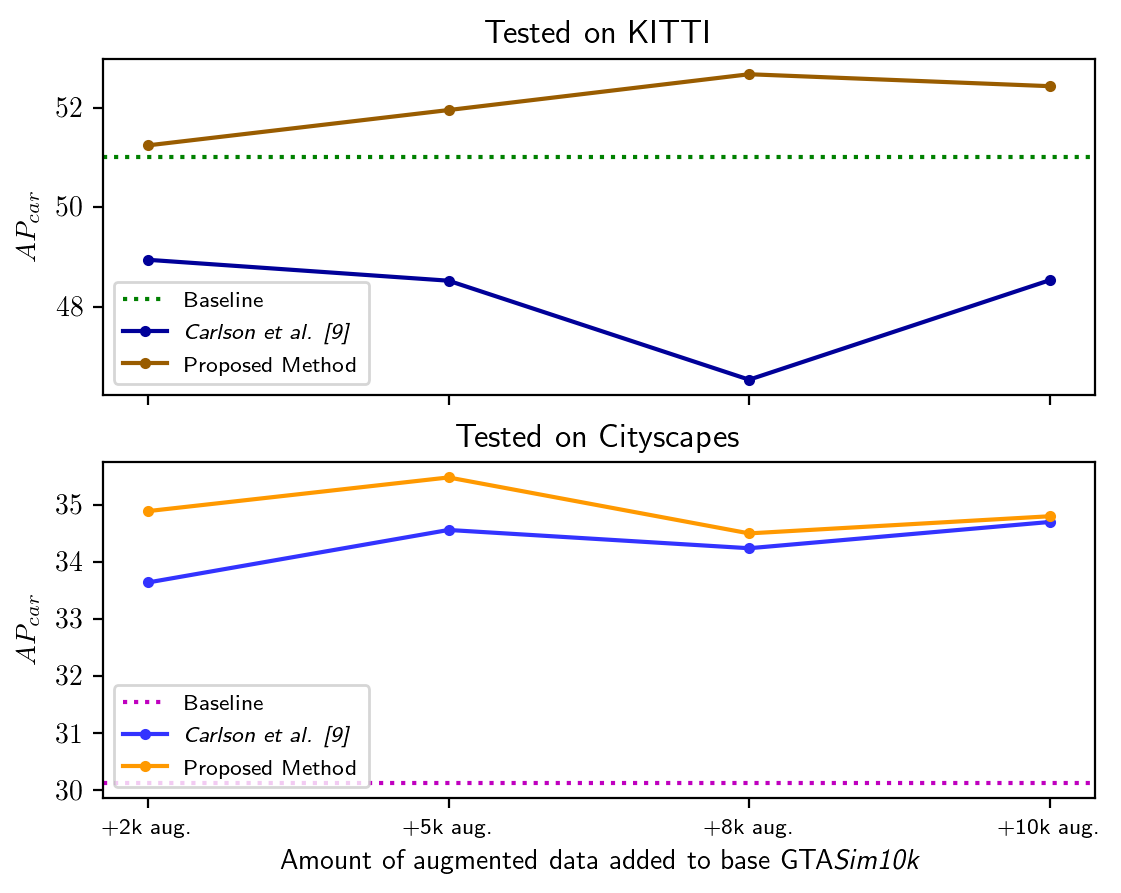}
\end{center}
\caption{Results of the learned sensor effect augmentations on Faster R-CNN object detection performance. Note that higher performance can be achieved using smaller synthetic datasets augmented with the proposed method for both KITTI and Cityscapes.}
\label{fig:numaug_results}
\end{figure}
To evaluate the impact of Sensor Transfer on the number of synthetic training images required for maximal object detection performance, we trained Faster R-CNNs on datasets comprised of the 10k unagumented GTA\textit{Sim10k} images combined with either 2k augmented images, 5k augmented images, 8k augmented images, or 10k augmented images. Figure~\ref{fig:numaug_results} captures the effect of increasing number of augmentations on Faster R-CNN performance. We see that, when compared to the Sensor Transfer domain randomization method, fewer training images are required when using Sensor Transfer augmentation for both GTA\textit{Sim10k}$\rightarrow$KITTI and GTA\textit{Sim10k}$\rightarrow$Cityscapes. 
Our results indicate that learning the augmentation parameters allows us to train on significantly smaller datasets without compromising performance. This demonstrates that we are more efficiently modeling salient visual information than domain randomization.
Interestingly, the Sensor Effect Domain Randomization method does worse than baseline across all levels of augmentation when tested on KITTI. We expect that this is because human-chosen set of parameter ranges, which are shown in the bottom row of Table~\ref{table:ST-hyperparameters}, do not generalize well when adapting GTA~\textit{Sim10k} to KITTI even though they may generate visually realistic images. One reason for this is that the visually realistic parameter ranges selected in ~\cite{carlson2018modeling} where chosen using a GTA dataset of all daytime images, whereas  GTA\textit{Sim10k} contains an even representation of daytime and nighttime images. This further demonstrates the importance of learning the sensor effect parameter distributions constrained by how they affect the styles of both the real and synthetic image datasets. 


\section{Discussion and Conclusions}
\label{sec:Discussion and Conclusions}
\label{sec:concl}
In general, our results show that the proposed Sensor Transfer Network reduces the synthetic to real domain gap more effectively and more efficiently than domain randomization. 
Future work includes increasingly the complexity and realism of the Sensor Transfer augmentation pipeline by modeling other, different sensor effects, as well as implementing models that better capture the pixel statistics of real images, such as motion or defocus blur. Other avenues include investigating the impact of task performance and problem space on the sensor effect parameter selection, and evaluating how the proposed method impacts performance for training synthetic datasets rendered with various levels of photorealism. 



\begin{thebibliography}{10}
\providecommand{\url}[1]{#1}
\csname url@samestyle\endcsname
\providecommand{\newblock}{\relax}
\providecommand{\bibinfo}[2]{#2}
\providecommand{\BIBentrySTDinterwordspacing}{\spaceskip=0pt\relax}
\providecommand{\BIBentryALTinterwordstretchfactor}{4}
\providecommand{\BIBentryALTinterwordspacing}{\spaceskip=\fontdimen2\font plus
\BIBentryALTinterwordstretchfactor\fontdimen3\font minus
  \fontdimen4\font\relax}
\providecommand{\BIBforeignlanguage}[2]{{%
\expandafter\ifx\csname l@#1\endcsname\relax
\typeout{** WARNING: IEEEtran.bst: No hyphenation pattern has been}%
\typeout{** loaded for the language `#1'. Using the pattern for}%
\typeout{** the default language instead.}%
\else
\language=\csname l@#1\endcsname
\fi
#2}}
\providecommand{\BIBdecl}{\relax}
\BIBdecl

\bibitem{kittiobject}
A.~Geiger, P.~Lenz, and R.~Urtasun, ``Are we ready for autonomous driving? the
  kitti vision benchmark suite,'' in \emph{Computer Vision and Pattern
  Recognition (CVPR), 2012 IEEE Conference on}.\hskip 1em plus 0.5em minus
  0.4em\relax IEEE, 2012, pp. 3354--3361.

\bibitem{KITTI13}
J.~Fritsch, T.~Kuehnl, and A.~Geiger, ``A new performance measure and
  evaluation benchmark for road detection algorithms,'' in \emph{International
  Conference on Intelligent Transportation Systems (ITSC)}, 2013.

\bibitem{Cityscapes}
M.~Cordts, M.~Omran, S.~Ramos, T.~Rehfeld, M.~Enzweiler, R.~Benenson,
  U.~Franke, S.~Roth, and B.~Schiele, ``The cityscapes dataset for semantic
  urban scene understanding,'' in \emph{Proceedings of the IEEE Conference on
  Computer Vision and Pattern Recognition}, 2016, pp. 3213--3223.

\bibitem{Richter_2016_ECCV}
S.~R. Richter, V.~Vineet, S.~Roth, and V.~Koltun, ``Playing for data: {G}round
  truth from computer games,'' in \emph{European Conference on Computer Vision
  (ECCV)}, ser. LNCS, B.~Leibe, J.~Matas, N.~Sebe, and M.~Welling, Eds., vol.
  9906.\hskip 1em plus 0.5em minus 0.4em\relax Springer International
  Publishing, 2016, pp. 102--118.

\bibitem{johnson2017driving}
M.~Johnson-Roberson, C.~Barto, R.~Mehta, S.~N. Sridhar, K.~Rosaen, and
  R.~Vasudevan, ``Driving in the matrix: Can virtual worlds replace
  human-generated annotations for real world tasks?'' in \emph{Robotics and
  Automation (ICRA), 2017 IEEE International Conference on}.\hskip 1em plus
  0.5em minus 0.4em\relax IEEE, 2017, pp. 746--753.

\bibitem{zhang2017image}
H.~Zhang, V.~Sindagi, and V.~M. Patel, ``Image de-raining using a conditional
  generative adversarial network,'' \emph{arXiv preprint arXiv:1701.05957},
  2017.

\bibitem{veeravasarapu2017adversarially}
V.~Veeravasarapu, C.~Rothkopf, and R.~Visvanathan, ``Adversarially tuned scene
  generation,'' \emph{arXiv preprint arXiv:1701.00405}, 2017.

\bibitem{sakaridis2018model}
C.~Sakaridis, D.~Dai, S.~Hecker, and L.~Van~Gool, ``Model adaptation with
  synthetic and real data for semantic dense foggy scene understanding,''
  \emph{arXiv preprint arXiv:1808.01265}, 2018.

\bibitem{carlson2018modeling}
A.~Carlson, K.~A. Skinner, and M.~Johnson-Roberson, ``Modeling camera effects
  to improve deep vision for real and synthetic data,'' \emph{arXiv preprint
  arXiv:1803.07721}, 2018.

\bibitem{grossberg2004modelingCRF}
M.~D. Grossberg and S.~K. Nayar, ``Modeling the space of camera response
  functions,'' vol.~26, no.~10.\hskip 1em plus 0.5em minus 0.4em\relax IEEE,
  2004, pp. 1272--1282.

\bibitem{couzinie2013learning_modelingblur}
F.~Couzinie-Devy, J.~Sun, K.~Alahari, and J.~Ponce, ``Learning to estimate and
  remove non-uniform image blur,'' in \emph{Proceedings of the IEEE Conference
  on Computer Vision and Pattern Recognition}, 2013, pp. 1075--1082.

\bibitem{foi2008practical}
A.~Foi, M.~Trimeche, V.~Katkovnik, and K.~Egiazarian, ``Practical
  poissonian-gaussian noise modeling and fitting for single-image raw-data,''
  vol.~17, no.~10.\hskip 1em plus 0.5em minus 0.4em\relax IEEE, 2008, pp.
  1737--1754.

\bibitem{andreopoulos2012sensor}
A.~Andreopoulos and J.~K. Tsotsos, ``On sensor bias in experimental methods for
  comparing interest-point, saliency, and recognition algorithms,'' vol.~34,
  no.~1.\hskip 1em plus 0.5em minus 0.4em\relax IEEE, 2012, pp. 110--126.

\bibitem{song2015sun}
S.~Song, S.~P. Lichtenberg, and J.~Xiao, ``Sun rgb-d: A rgb-d scene
  understanding benchmark suite,'' in \emph{Proceedings of the IEEE conference
  on computer vision and pattern recognition}, 2015, pp. 567--576.

\bibitem{dodge2016understanding}
S.~Dodge and L.~Karam, ``Understanding how image quality affects deep neural
  networks,'' in \emph{Quality of Multimedia Experience (QoMEX), 2016 Eighth
  International Conference on}.\hskip 1em plus 0.5em minus 0.4em\relax IEEE,
  2016, pp. 1--6.

\bibitem{doersch2015unsupervised}
C.~Doersch, A.~Gupta, and A.~A. Efros, ``Unsupervised visual representation
  learning by context prediction,'' in \emph{Proceedings of the IEEE
  International Conference on Computer Vision}, 2015, pp. 1422--1430.

\bibitem{zhang2017physically}
Y.~Zhang, S.~Song, E.~Yumer, M.~Savva, J.-Y. Lee, H.~Jin, and T.~Funkhouser,
  ``Physically-based rendering for indoor scene understanding using
  convolutional neural networks,'' in \emph{2017 IEEE Conference on Computer
  Vision and Pattern Recognition (CVPR)}.\hskip 1em plus 0.5em minus
  0.4em\relax IEEE, 2017, pp. 5057--5065.

\bibitem{tobin2017domain}
J.~Tobin, R.~Fong, A.~Ray, J.~Schneider, W.~Zaremba, and P.~Abbeel, ``Domain
  randomization for transferring deep neural networks from simulation to the
  real world,'' in \emph{Intelligent Robots and Systems (IROS), 2017 IEEE/RSJ
  International Conference on}.\hskip 1em plus 0.5em minus 0.4em\relax IEEE,
  2017, pp. 23--30.

\bibitem{tremblay2018training}
J.~Tremblay, A.~Prakash, D.~Acuna, M.~Brophy, V.~Jampani, C.~Anil, T.~To,
  E.~Cameracci, S.~Boochoon, and S.~Birchfield, ``Training deep networks with
  synthetic data: Bridging the reality gap by domain randomization,''
  \emph{arXiv preprint arXiv:1804.06516}, 2018.

\bibitem{paulin2014transformation}
M.~Paulin, J.~Revaud, Z.~Harchaoui, F.~Perronnin, and C.~Schmid,
  ``Transformation pursuit for image classification,'' in \emph{Computer Vision
  and Pattern Recognition (CVPR), 2014 IEEE Conference on}.\hskip 1em plus
  0.5em minus 0.4em\relax IEEE, 2014, pp. 3646--3653.

\bibitem{cubuk2018autoaugment}
E.~D. Cubuk, B.~Zoph, D.~Mane, V.~Vasudevan, and Q.~V. Le, ``Autoaugment:
  Learning augmentation policies from data,'' \emph{arXiv preprint
  arXiv:1805.09501}, 2018.

\bibitem{lemley2017smart}
J.~Lemley, S.~Bazrafkan, and P.~Corcoran, ``Smart augmentation learning an
  optimal data augmentation strategy.'' \emph{IEEE Access}, vol.~5, pp.
  5858--5869, 2017.

\bibitem{CycleGAN2017}
J.-Y. Zhu, T.~Park, P.~Isola, and A.~A. Efros, ``Unpaired image-to-image
  translation using cycle-consistent adversarial networks,'' \emph{arXiv
  preprint arXiv:1703.10593}, 2017.

\bibitem{johnson2016perceptual}
J.~Johnson, A.~Alahi, and L.~Fei-Fei, ``Perceptual losses for real-time style
  transfer and super-resolution,'' in \emph{European Conference on Computer
  Vision}.\hskip 1em plus 0.5em minus 0.4em\relax Springer, 2016, pp. 694--711.

\bibitem{zhu2016generative}
J.-Y. Zhu, P.~Kr{\"a}henb{\"u}hl, E.~Shechtman, and A.~A. Efros, ``Generative
  visual manipulation on the natural image manifold,'' in \emph{European
  Conference on Computer Vision}.\hskip 1em plus 0.5em minus 0.4em\relax
  Springer, 2016, pp. 597--613.

\bibitem{isola2017image}
P.~Isola, J.-Y. Zhu, T.~Zhou, and A.~A. Efros, ``Image-to-image translation
  with conditional adversarial networks,'' \emph{arXiv preprint}, 2017.

\bibitem{huang2018multimodal}
X.~Huang, M.-Y. Liu, S.~Belongie, and J.~Kautz, ``Multimodal unsupervised
  image-to-image translation,'' \emph{arXiv preprint arXiv:1804.04732}, 2018.

\bibitem{liu2017unsupervised}
M.-Y. Liu, T.~Breuel, and J.~Kautz, ``Unsupervised image-to-image translation
  networks,'' in \emph{Advances in Neural Information Processing Systems},
  2017, pp. 700--708.

\bibitem{dundar2018domain}
A.~Dundar, M.-Y. Liu, T.-C. Wang, J.~Zedlewski, and J.~Kautz, ``Domain
  stylization: A strong, simple baseline for synthetic to real image domain
  adaptation,'' \emph{arXiv preprint arXiv:1807.09384}, 2018.

\bibitem{sixt2016rendergan}
L.~Sixt, B.~Wild, and T.~Landgraf, ``Rendergan: Generating realistic labeled
  data,'' \emph{arXiv preprint arXiv:1611.01331}, 2016.

\bibitem{simgan}
A.~Shrivastava, T.~Pfister, O.~Tuzel, J.~Susskind, W.~Wang, and R.~Webb,
  ``Learning from simulated and unsupervised images through adversarial
  training,'' \emph{arXiv preprint arXiv:1612.07828}, 2016.

\bibitem{huangexpecting}
S.~Huang, D.~Ramanan, undefined, undefined, undefined, and undefined,
  ``Expecting the unexpected: Training detectors for unusual pedestrians with
  adversarial imposters,'' \emph{2017 IEEE Conference on Computer Vision and
  Pattern Recognition (CVPR)}, vol.~00, pp. 4664--4673, 2017.

\bibitem{Kanan2012}
C.~Kanan and G.~W. Cottrell, ``Color-to-grayscale: does the method matter in
  image recognition?'' \emph{PloS one}, vol.~7, no.~1, p. e29740, 2012.

\bibitem{diamond2017dirty}
S.~Diamond, V.~Sitzmann, S.~Boyd, G.~Wetzstein, and F.~Heide, ``Dirty pixels:
  Optimizing image classification architectures for raw sensor data,'' 2017.

\bibitem{cheong2015fast}
H.~Cheong, E.~Chae, E.~Lee, G.~Jo, and J.~Paik, ``Fast image restoration for
  spatially varying defocus blur of imaging sensor,'' \emph{Sensors}, vol.~15,
  no.~1, pp. 880--898, 2015.

\bibitem{exp1}
S.~A. Bhukhanwala and T.~V. Ramabadran, ``Automated global enhancement of
  digitized photographs,'' \emph{IEEE Transactions on Consumer Electronics},
  vol.~40, no.~1, pp. 1--10, Feb 1994.

\bibitem{exp2}
G.~Messina, A.~Castorina, S.~Battiato, and A.~Bosco, ``Image quality
  improvement by adaptive exposure correction techniques,'' in \emph{Multimedia
  and Expo, 2003. ICME '03. Proceedings. 2003 International Conference on},
  vol.~1, July 2003, pp. I--549--52 vol.1.

\bibitem{hunter1948accuracyLAB}
R.~S. Hunter, ``Accuracy, precision, and stability of new photoelectric
  color-difference meter,'' in \emph{Journal of the Optical Society of
  America}, vol.~38, no.~12, 1948, pp. 1094--1094.

\bibitem{annadurai2007fundamentals}
S.~Annadurai, ``Fundamentals of digital image processing.''\hskip 1em plus
  0.5em minus 0.4em\relax Pearson Education India, 2007.

\bibitem{vgg16}
\BIBentryALTinterwordspacing
K.~Simonyan and A.~Zisserman, ``Very deep convolutional networks for
  large-scale image recognition,'' \emph{CoRR}, vol. abs/1409.1556, 2014.
  [Online]. Available: \url{http://arxiv.org/abs/1409.1556}
\BIBentrySTDinterwordspacing

\bibitem{deng2009imagenet}
J.~Deng, W.~Dong, R.~Socher, L.-J. Li, K.~Li, and L.~Fei-Fei, ``Imagenet: A
  large-scale hierarchical image database,'' in \emph{Computer Vision and
  Pattern Recognition, 2009. CVPR 2009. IEEE Conference on}.\hskip 1em plus
  0.5em minus 0.4em\relax Ieee, 2009, pp. 248--255.

\bibitem{ren2015fasterRCNN}
S.~Ren, K.~He, R.~Girshick, and J.~Sun, ``Faster r-cnn: Towards real-time
  object detection with region proposal networks,'' in \emph{Advances in neural
  information processing systems}, 2015, pp. 91--99.

\bibitem{pascal-voc-2012}
M.~Everingham, L.~Van~Gool, C.~K.~I. Williams, J.~Winn, and A.~Zisserman, ``The
  {PASCAL} {V}isual {O}bject {C}lasses {C}hallenge 2012 {(VOC2012)}
  {R}esults,''
  http://www.pascal-network.org/challenges/VOC/voc2012/workshop/index.html.

\end{thebibliography}
\end{document}